\documentclass[letterpaper]{article} 
\usepackage{aaai25}  
\usepackage{times}  
\usepackage{helvet}  
\usepackage{courier}  
\usepackage[hyphens]{url}  
\usepackage{graphicx} 
\usepackage{natbib}  
\usepackage{caption} 
\usepackage{algorithm}
\usepackage{algorithmic}

\newcommand{\ie}{{\emph{i.e.}}, }

\newcommand{\eg}{{\emph{e.g.}}, }
\usepackage{amsmath}

\usepackage{xcolor}
\usepackage{tabularx} 
\usepackage{multirow}
\usepackage{pbox}
\usepackage{graphicx}
{\begin{list}               
    {$\bullet$ \hfill}{
        \setlength{\leftmargin}{\parindent}
        \setlength{\parsep}{0.04\baselineskip}
        \setlength{\itemsep}{0.5\parsep}
        \setlength{\labelwidth}{\leftmargin}
        \setlength{\labelsep}{0em}}
    }
{\end{list}}

\providecommand{\eref}[1]{Eq. \eqref{#1}}  
\providecommand{\cref}[1]{Chapter~\ref{#1}}
\providecommand{\sref}[1]{Section~\ref{#1}}
\providecommand{\fref}[1]{Figure~\ref{#1}}
\providecommand{\tref}[1]{Table~\ref{#1}}

\providecommand{\norm}[1]{\lVert#1\rVert}

\renewcommand{\vec}[1]{\ensuremath{\boldsymbol{#1}}}
\providecommand{\mat}[1]{\ensuremath{\boldsymbol{#1}}}

\providecommand{\calB}{\mathcal{B}}
\providecommand{\calC}{\mathcal{C}}
\providecommand{\calD}{\mathcal{D}}

\providecommand{\calH}{\mathcal{H}}
\providecommand{\calI}{\mathcal{I}}

\providecommand{\calL}{\mathcal{L}}

\providecommand{\calZ}{\mathcal{Z}}

\providecommand{\mA}{\mat{A}}

\providecommand{\mF}{\mat{F}}
\providecommand{\mG}{\mat{G}}

\providecommand{\mI}{\mat{I}}

\providecommand{\mM}{\mat{M}}

\providecommand{\mO}{\mat{O}}
\providecommand{\mP}{\mat{P}}

\providecommand{\mS}{\mat{S}}

\providecommand{\mW}{\mat{W}}

\providecommand{\vq}{\vec{q}}

\providecommand{\vy}{\vec{y}}
\providecommand{\vz}{\vec{z}}

\usepackage{booktabs}       
\usepackage{amsfonts}       
\usepackage{url}            
\usepackage[T1]{fontenc}

\usepackage{newfloat}
\usepackage{listings}
\DeclareCaptionStyle{ruled}{labelfont=normalfont,labelsep=colon,strut=off} 
\floatstyle{ruled}
\newfloat{listing}{tb}{lst}{}
\floatname{listing}{Listing}
\title{Generalized Class Discovery in Instance Segmentation}
\author {
    Cuong Manh Hoang\textsuperscript{\rm 1},
    Yeejin Lee\textsuperscript{\rm 1},
    Byeongkeun Kang\textsuperscript{\rm 2}\thanks{Corresponding author.}
}
\affiliations {
    \textsuperscript{\rm 1}Seoul National University of Science and Technology, Republic of Korea\\
     \textsuperscript{\rm 2}Chung-Ang University, Republic of Korea\\
    \{cuonghoang, yeejinlee\}@seoultech.ac.kr, byeongkeunkang@cau.ac.kr
}

\usepackage{bibentry}

\begin{document}

\maketitle

\begin{abstract}
This work addresses the task of generalized class discovery (GCD) in instance segmentation. The goal is to discover novel classes and obtain a model capable of segmenting instances of both known and novel categories, given labeled and unlabeled data. Since the real world contains numerous objects with long-tailed distributions, the instance distribution for each class is inherently imbalanced. To address the imbalanced distributions, we propose an instance-wise temperature assignment (ITA) method for contrastive learning and class-wise reliability criteria for pseudo-labels. The ITA method relaxes instance discrimination for samples belonging to head classes to enhance GCD. The reliability criteria are to avoid excluding most pseudo-labels for tail classes when training an instance segmentation network using pseudo-labels from GCD. Additionally, we propose dynamically adjusting the criteria to leverage diverse samples in the early stages while relying only on reliable pseudo-labels in the later stages. We also introduce an efficient soft attention module to encode object-specific representations for GCD. Finally, we evaluate our proposed method by conducting experiments on two settings: COCO$_{half}$ + LVIS and LVIS + Visual Genome. The experimental results demonstrate that the proposed method outperforms previous state-of-the-art methods.
\end{abstract}

%



\section{Introduction}
\label{sec:intro}
While supervised instance segmentation methods~\cite{he2017mask,cheng2022masked,jain2023oneformer} have achieved impressive performance, they require large-scale datasets with expensive human annotations. To reduce annotation costs, researchers have investigated semi-supervised learning (SSL) methods~\cite{bellver2019budget,yang2022bias, Berrada2024Guided} that utilize unlabeled images along with small-scale labeled data, as well as weakly-supervised methods~\cite{Lan2021DiscoBox} relying on weak annotations. However, all these methods rely on the closed-world assumption and can recognize only the objects belonging to the classes (\ie known classes) in the labeled dataset.


To address this limitation, researchers have introduced novel category discovery (NCD)~\cite{han2019learning}. Unlike SSL, where the unlabeled images contain only known classes, NCD assumes that the unlabeled data include novel categories. Recently, generalized (novel) category discovery (GCD)~\cite{vaze2022generalized, cao2021open} was introduced, further relaxing the assumptions on the unlabeled data. It assumes that the unlabeled data may contain both known and novel classes, making the problem more challenging and realistic. Given labeled and unlabeled data, GCD aims to train a model capable of recognizing both the known classes (\eg person and car) in the labeled data and the novel categories (\eg unknown$_1$ and unknown$_2$) discovered from the unlabeled data.


Most previous works have investigated GCD for curated and balanced image classification datasets~\cite{vaze2022generalized, An2023Generalized, Zhang2023PromptCAL, Pu2023Dynamic, wen2023parametric}. Recently, researchers have explored GCD for image classification on imbalanced datasets~\cite{Bai2023Towards, li2023imbagcd, li2023generalized}, semantic segmentation~\cite{zhao2022novel}, 3D point cloud semantic segmentation~\cite{riz2023novel}, and instance segmentation~\cite{fomenko2022learning, weng2021unsupervised}. Most of these works leverage semi-supervised contrastive learning and pseudo-label generation regardless of tasks. 



In this work, we also investigate GCD in instance segmentation. It is worth noting that because the real world contains numerous objects with long-tailed distributions, the instance distribution for each class in instance segmentation datasets is inherently imbalanced. For example, a `car' appears more frequently than an `ashtray'. (1) To address this imbalanced distribution, we propose an instance-wise temperature assignment method for contrastive learning. While typical contrastive learning losses treat samples from head and tail classes equally, we aim to emphasize group-wise discrimination for head class samples while focusing on instance-wise discrimination for tail class samples, inspired by~\cite{kukleva2023temperature}. (2) Although relying on reliable pseudo-labels is important~\cite{yang2022st++}, applying fixed and global reliability criteria to imbalanced data tends to exclude most pseudo-labels for tail classes. Therefore, we propose to utilize class-wise reliability criteria to apply varying thresholds for head and tail classes. Additionally, we dynamically adjust the reliability criteria throughout training to leverage diverse samples in the early stages while focusing on reliable pseudo-labels in later stages. (3) Finally, we introduce an efficient soft attention module based on spatial pooling and depth reduction to effectively encode representations for target instances while suppressing those from background or adjacent objects.


The contributions of this paper are as follows: (1) We propose an instance-wise temperature assignment method for semi-supervised contrastive learning in GCD to enhance the separability of classes in long-tailed distributions. (2) We introduce a reliability-based dynamic learning method for training with pseudo-labels to apply different reliability criteria to each class based on its tailness. Additionally, we adjust these criteria during training to rely on strictly reliable pseudo-labels in the later stages while using diverse data in the early stages. (3) We propose an efficient soft attention module for encoding object-specific representations for GCD. (4) We validate the effectiveness of the proposed method by discovering novel classes and training an instance segmentation network using labeled and unlabeled data.

\section{Related Works}
\noindent \textbf{Generalized Class Discovery in Image Classification}. 
\cite{vaze2022generalized} introduced GCD, which aims to categorize unlabeled images given both labeled and unlabeled data. They trained an embedding network using unsupervised contrastive learning on all the data and supervised contrastive learning on the labeled data. They then applied semi-supervised $k$-means clustering to assign class or cluster labels to the unlabeled images. Additionally, they proposed a method for estimating the number of novel classes.


\cite{An2023Generalized} proposed DPN, which utilizes two sets of category-wise prototypes: one for labeled data and the other for unlabeled images based on $k$-means clustering. For clustering, they assumed prior knowledge of the total number of categories. They discovered novel classes in the unlabeled data by applying the Hungarian algorithm~\cite{kuhn1955hungarian} to the two sets. \cite{Zhang2023PromptCAL} presented PromptCAL, which uses an affinity graph to generate pseudo-labels for unlabeled data. \cite{Pu2023Dynamic} introduced the DCCL framework, which employs a hyperparameter-free clustering algorithm~\cite{Martin2008Maps} to generate pseudo-labels in the absence of ground-truth cluster numbers. \cite{wen2023parametric} proposed SimGCD, a one-stage framework that replaces the separate semi-supervised clustering in~\cite{vaze2022generalized} with a jointly trainable parametric classifier. They analyzed the problem of using unreliable pseudo-labels for training a parametric classifier and proposed using soft pseudo-labels.


Recently, \cite{Bai2023Towards} introduced GCD for long-tailed datasets to address imbalanced distributions in real-world scenarios. They proposed the BaCon framework, which includes a pseudo-labeling branch and a contrastive learning branch. To handle imbalanced distributions, they estimated data distribution using $k$-means clustering and the Hungarian algorithm~\cite{kuhn1955hungarian}. Concurrently, \cite{li2023imbagcd} proposed ImbaGCD, which estimates class prior distributions assuming unknown classes are usually tail classes. They then generated pseudo-labels for unlabeled images using the estimated prior distribution and the Sinkhorn-Knopp algorithm~\cite{Cuturi2013Sinkhorn}. Later, \cite{li2023generalized} replaced the expectation–maximization (EM) and Sinkhorn-Knopp~\cite{Cuturi2013Sinkhorn} algorithms in~\cite{li2023imbagcd} with cross-entropy-based regularization losses to reduce computational costs.

\noindent \textbf{Class Discovery in Segmentation and Detection}. 
\cite{zhao2022novel} introduced NCD in semantic segmentation. They proposed to find novel salient object regions using a saliency model and a segmentation network trained on labeled data. They then applied clustering to these object regions to obtain pseudo-labels. Finally, they trained a segmentation network using the labeled data and the unlabeled images with clean pseudo-labels from clustering and online pseudo-labels. The clean pseudo-labels were dynamically assigned based on entropy ranking. \cite{riz2023novel} extended the method from~\cite{zhao2022novel} to 3D point cloud semantic segmentation. Specifically, they utilized online clustering and exploited uncertainty quantification to generate pseudo-labels.

In instance segmentation, \cite{weng2021unsupervised} investigated unsupervised long-tail category discovery. They initially employed a class-agnostic mask proposal network to obtain masks for all objects. They then trained an embedding network using self-supervised triplet losses. Finally, they applied hyperbolic $k$-means clustering to discover novel categories. \cite{fomenko2022learning} introduced novel class discovery and localization, which can be viewed as GCD in object detection and instance segmentation. They first trained a Faster R-CNN~\cite{ren2015faster} or Mask R-CNN~\cite{he2017mask} using the labeled data and froze the network except for its classification head. Then, they applied the frozen network to unlabeled and labeled images to obtain region proposals. Subsequently, they expanded the classification head to incorporate new classes and trained it using pseudo-labels generated by online clustering based on the Sinkhorn-Knopp algorithm~\cite{Cuturi2013Sinkhorn}.

\begin{figure*}[!t] \begin{center}
\begin{minipage}{1\linewidth}
\centerline{\includegraphics[scale=0.55]{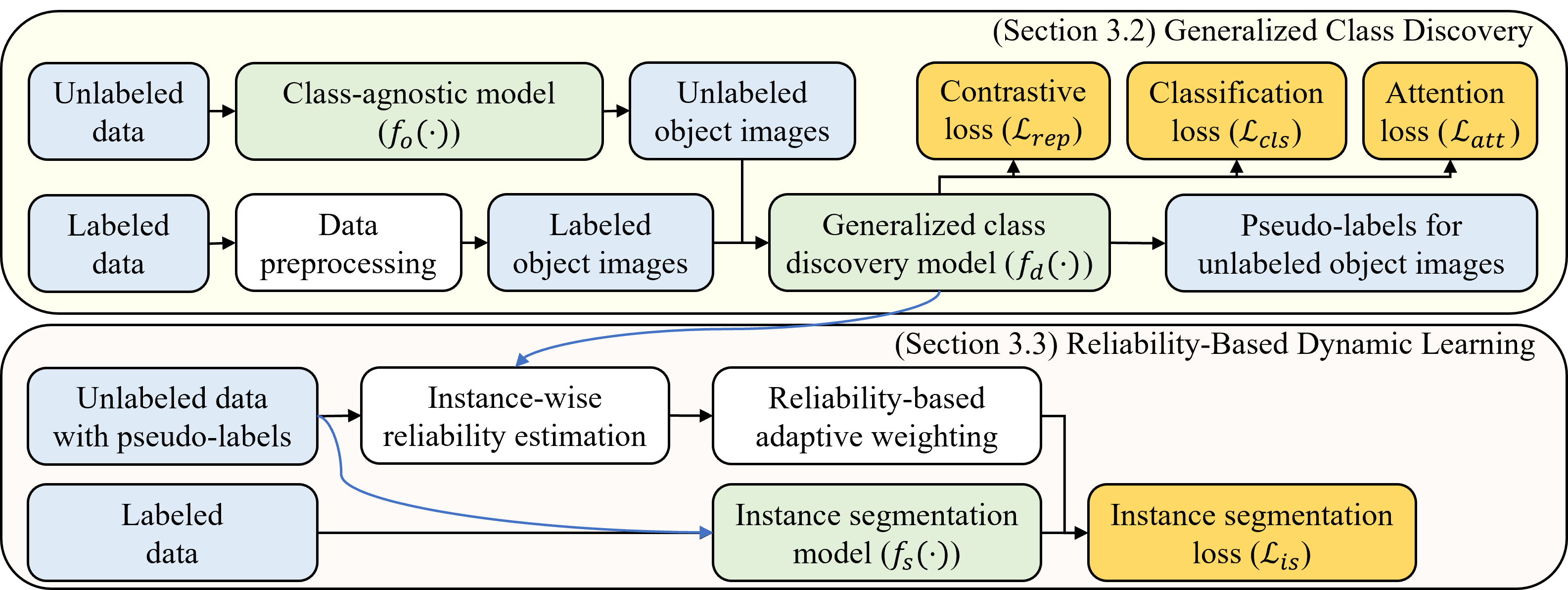}}
\end{minipage}
   \caption{Overview of the proposed framework during training. We first train a class-agnostic instance segmentation network $f_o(\cdot)$ and apply it to unlabeled images to generate class-agnostic instance masks. Then, we train the GCD model $f_d(\cdot)$ using unlabeled object images and labeled object images to discover novel classes in the unlabeled data. Finally, we train an instance segmentation network $f_s(\cdot)$ using the labeled data and the unlabeled images with pseudo-labels.}
\label{fig:overview_training}
\end{center}\end{figure*}

\section{Proposed Method}
\label{sec:method}
We first define the GCD problem in instance segmentation, which aims to discover novel classes and learn to segment instances of both known and novel categories, given labeled and unlabeled data. We then present our GCD method in~\sref{sec:class_discovery} and the method for training an instance segmentation network in~\sref{sec:dynamic_learning}. An overview of the proposed framework during training is illustrated in~\fref{fig:overview_training}.


\subsection{Preliminaries}
\label{sec:prelim}
\noindent \textbf{Problem Formulation}.
We are given a labeled dataset $\calD^l$ and an unlabeled dataset $\calD^u$. $\calD^l$ contains images $\{\mI^l\}$ along with instance-wise class and mask labels ($\{\vy^l\}$, $\{\mM^l\}$) for known classes $\calC^k$, while $\calD^u$ comprises only images $\{\mI^u\}$. Given $\calD^l$ and $\calD^u$, GCD in instance segmentation aims to discover novel categories $\calC^n$ (\ie $\calC^k \cap \calC^n = \emptyset$) and to obtain a model capable of segmenting instances of both the known and novel classes $\calC = \calC^k \cup \calC^n$. Hence, during inference, the network is expected to segment instances of known classes (\eg person and car) as well as novel categories (\eg $\text{unknown}_1$ and $\text{unknown}_2$) given an image $\mI$. The images in $\calD^l$ and $\calD^u$ may contain instances of both the known and novel classes.

\vspace{1mm}
\noindent \textbf{Contrastive Learning for GCD}. 
\cite{vaze2022generalized} introduced a contrastive learning (CL) method for GCD in balanced image classification datasets. They first pre-trained a backbone with DINO~\cite{dino2021} on the ImageNet dataset~\cite{imagenet2015} without labels. Subsequently, they fine-tuned the backbone and a projection head using supervised CL on the labeled data and unsupervised CL on both the labeled and unlabeled data. 


Following~\cite{wen2023parametric, vaze2022generalized}, our GCD model $f_d(\cdot)$ consists of a backbone $b(\cdot)$ and a projection head $g(\cdot)$. We utilize an MLP for $g(\cdot)$ and a ResNet-50 backbone~\cite{resnet} for $b(\cdot)$, which is pre-trained on the unlabeled ImageNet dataset~\cite{imagenet2015} using DINO~\cite{dino2021}. Additionally, we employ a momentum encoder $f'_d(\cdot)$ whose parameters are momentum-based moving averages of the parameters of $f_d(\cdot)$ during training, following MoCo~\cite{he2020momentum}. We also use a queue to store the embeddings from $f'_d(\cdot)$ for the samples of both the previous and current mini-batches.




Formally, given an image, we generate two views (random augmentations) $\mI_i$ and $\mI'_i$. We then encode them using $f_d(\cdot)$ and $f'_d(\cdot)$ to obtain $\vz_i = f_d(\mI_i) = g(b(\mI_i))$ and $\vz'_i = f'_d(\mI'_i)$, respectively. We store $\vz'_i$ from the samples in the previous and current mini-batches in the queue $\calZ'$. In~\cite{vaze2022generalized}, the unsupervised contrastive loss $\calL^u_{rep}$ and supervised contrastive loss $\calL^s_{rep}$ are computed as follows:
\begin{equation}
\begin{split}
& \calL^u_{rep}:= -\log \frac{\exp(\vz_i^\mathsf{T} \vz'_i/\tau)}{\sum_{\vz'_j \in \hat{\calZ}_i } \exp(\vz_i^\mathsf{T} \vz'_j/\tau)}, \\
& \calL^s_{rep} := - \frac{1}{|\calZ^p_i|}\sum_{\vz'_k \in \calZ^p_i} \log \frac{\exp(\vz_i^\mathsf{T} \vz'_k/\tau)}{\sum_{\vz'_j \in \hat{\calZ}_i } \exp(\vz_i^\mathsf{T} \vz'_j/\tau)}
\label{eqn:cont_loss_prelim}
\end{split}
\end{equation}
where $\hat{\calZ}_i$ represents the set $\calZ'$ excluding $\vz'_i$ (\ie $\hat{\calZ}_i = \calZ' \setminus \vz'_i$); $\calZ^p_i$ denotes the subset of $\calZ'$ containing the representations that belong to the same class as $\mI_i$; $\tau$ is a temperature hyperparameter; and $|\cdot|$ denotes the number of samples in the set.

\subsection{Generalized Class Discovery}
\label{sec:class_discovery}
Given $\calD^l$ and $\calD^u$, we aim to discover novel categories in $\calD^u$ using the knowledge from $\calD^l$. To achieve this, we first train an instance segmentation network using $\calD^l$ and $\calD^u$ to generate class-agnostic instance masks $\mM^u$ for all objects in $\calD^u$. Subsequently, we crop the unlabeled images using $\mM^u$ and the labeled images using the ground-truth masks $\mM^l$. Then, we train a GCD model using the cropped unlabeled images $\calI^u_o$ and the cropped labeled images $\calI^l_o$ with class labels $\vy^l$ to generate pseudo-class labels $\vy^u$ for $\calI^u_o$.



\vspace{1mm}
\noindent \textbf{Class-Agnostic Instance Mask Generation}.
Similar to~\cite{fomenko2022learning}, we first train an instance segmentation network $f_o(\cdot)$ to obtain class-agnostic instance masks for both known and unseen classes $\calC = \calC^k \cup \calC^n$. We train a class-agnostic instance segmentation network, the Generic Grouping Network (GGN) from~\cite{wang2022open}, using both $\calD^l$ and $\calD^u$. We experimentally demonstrate that our GCD method is robust when applied to other class-agnostic instance segmentation methods.



Once the training terminates, we apply $f_o(\cdot)$ to the images $\mI^u \in \calD^u$ to obtain instance masks $\mM^u$. We then construct an unlabeled object image set $\calI^u_o$ by cropping the rectangular regions of $\mI^u$ based on $\mM^u$. Similarly, a labeled object image set $\calI^l_o$ is prepared by cropping $\mI^l$ in $\calD^l$ based on the mask labels $\mM^l$.







\vspace{1mm}
\noindent \textbf{Contrastive Learning for GCD in Instance Segmentation}. 
We propose a contrastive learning method for GCD in instance segmentation by modifying the losses in~\eref{eqn:cont_loss_prelim}. While these losses are designed for curated and balanced data, instances in typical instance segmentation datasets are naturally imbalanced (\ie certain objects appear more frequently than others). To address the long-tail distribution of instances for each class, we propose to adjust the temperature parameters in $\calL^u_{rep}$ and $\calL^s_{rep}$ for each instance based on its likelihood of belonging to head classes.

\begin{figure}[!t] 
\begin{minipage}{0.49\linewidth}
\centerline{\includegraphics[width=0.97\linewidth,height=0.07\textheight]{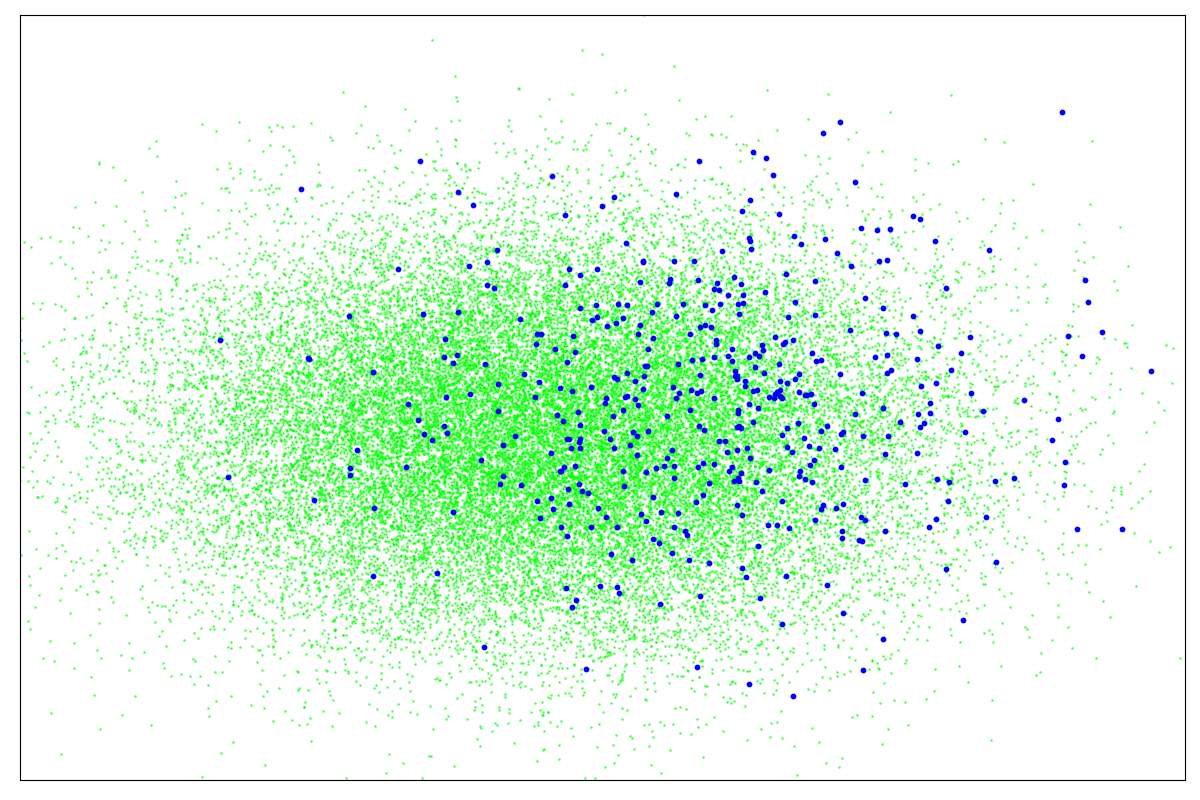}}
\end{minipage}
\begin{minipage}{0.49\linewidth}
\centerline{\includegraphics[width=0.97\linewidth,height=0.07\textheight]{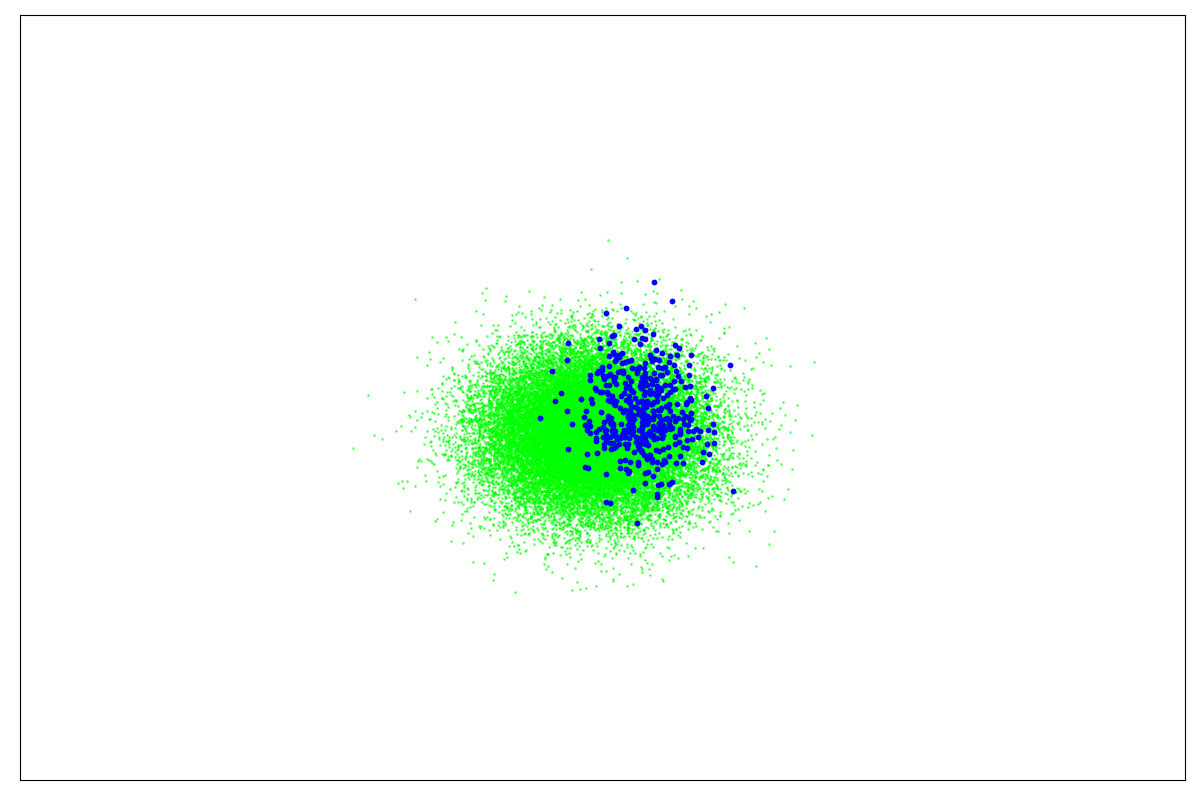}}
\end{minipage}

\begin{minipage}{0.49\linewidth}
\centerline{\footnotesize (a) $\tau=0.07$}
\end{minipage}
\begin{minipage}{0.49\linewidth}
\centerline{\footnotesize (b) $\tau=1$}
\end{minipage}

\begin{minipage}{0.49\linewidth}
\centerline{\includegraphics[width=0.97\linewidth,height=0.07\textheight]{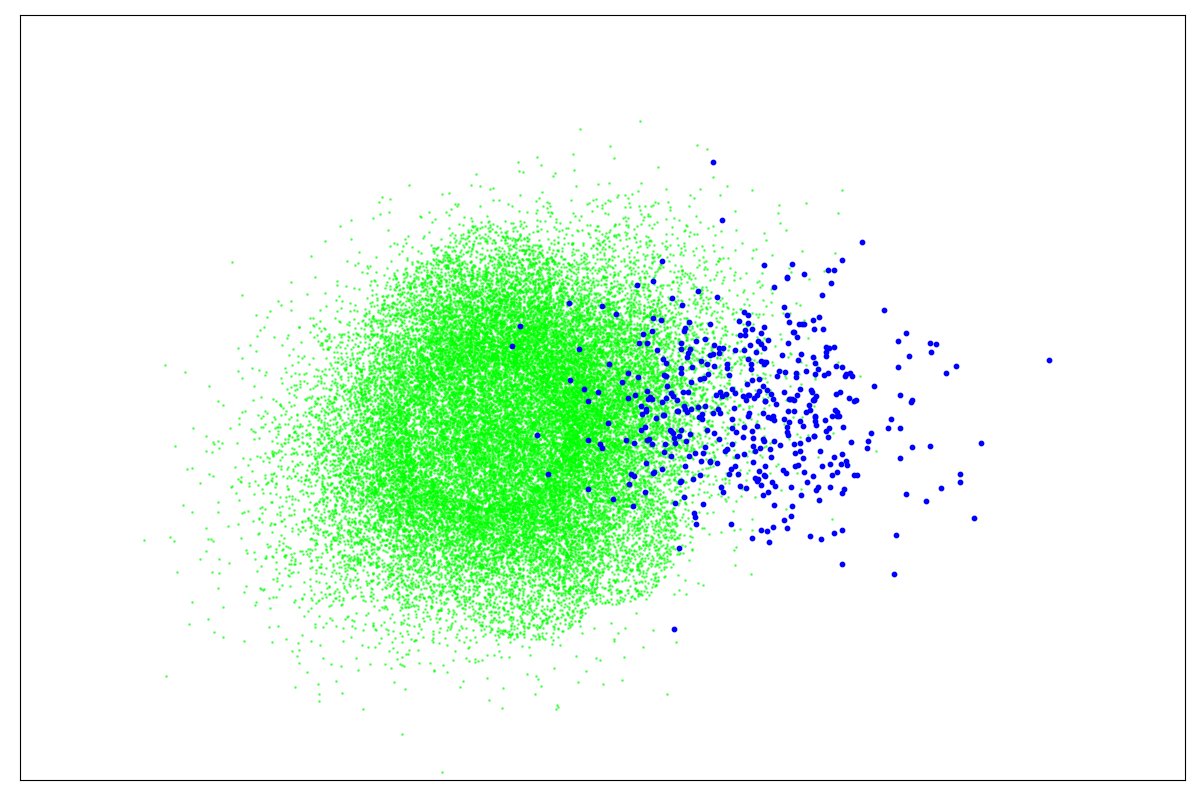}}
\end{minipage}
\begin{minipage}{0.49\linewidth}
\centerline{\includegraphics[width=0.97\linewidth,height=0.07\textheight]{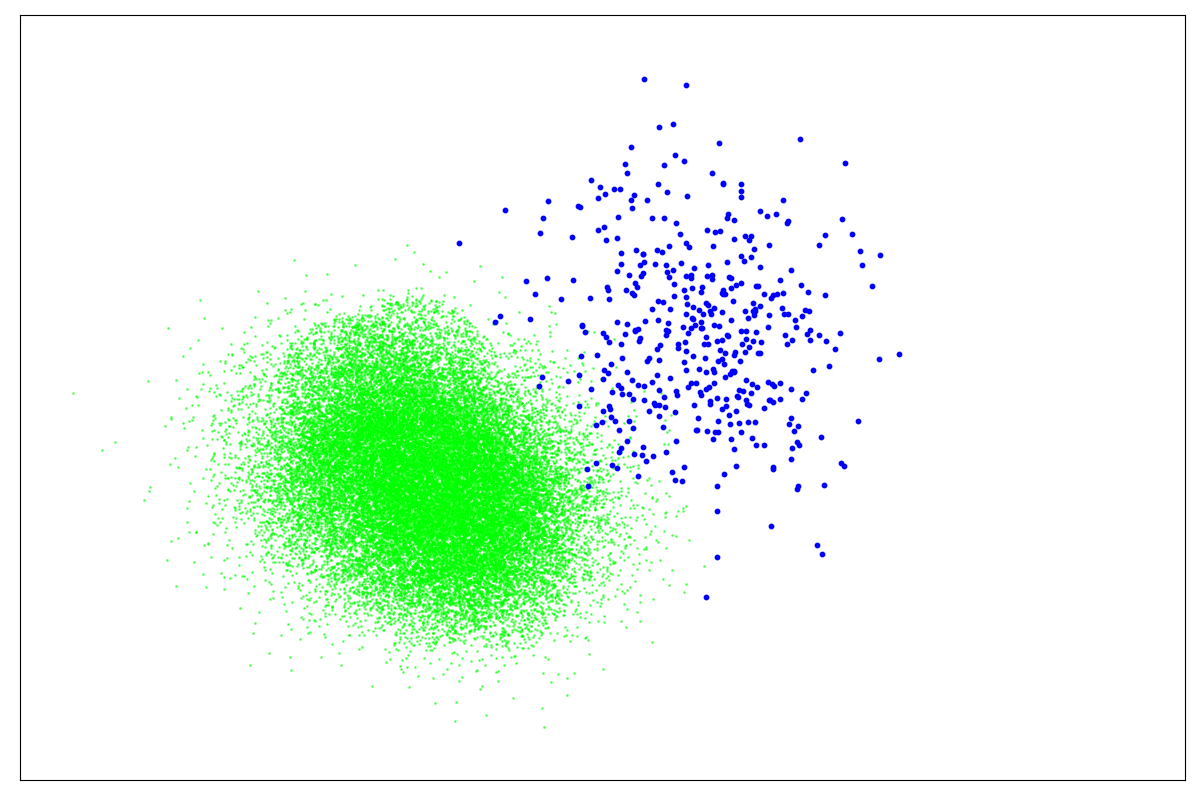}}
\end{minipage}

\begin{minipage}{0.49\linewidth}
\centerline{\footnotesize (c) TS}
\end{minipage}
\begin{minipage}{0.49\linewidth}
\centerline{\footnotesize (d) Ours (ITA)}
\end{minipage}
\caption{$t$-SNE visualization of two semantically close classes. Green: single head class (`book'); Blue: single tail class (`booklet').}
\label{fig:ITA}
\end{figure}

\fref{fig:ITA} visualizes $t$-SNE projections of two semantically similar classes: `book' and `booklet'. Previously, \cite{kukleva2023temperature} showed that a low temperature value tends to uniformly discriminate all instances while a high temperature value leads to group-wise discrimination, as shown in~\fref{fig:ITA} (a) and (b). Based on this, \cite{kukleva2023temperature} introduced a cosine temperature scheduling (TS) method to alternate between instance-wise and group-wise discrimination. In contrast, we propose to estimate the headness of each instance and assign high/low temperature values to instances belonging to head/tail class samples. \fref{fig:ITA} demonstrates the superiority of our instance-wise temperature assignment (ITA) method compared to static assignments and TS~\cite{kukleva2023temperature}.





In detail, we first compute a headness score $\hat{h}_i$ for each instance $\mI_i$ by estimating the density of the neighborhood of $\vz_i$ in the embedding space. A higher score $\hat{h}_i$ indicates a higher probability of $\mI_i$ belonging to a head class. We then apply a momentum update to the headness scores to enhance the robustness of the estimation. Specifically, $\hat{h}_i$ and the momentum-updated headness score $h_i$ at the $t$-th epoch are computed as follows:
\begin{equation}
\begin{split}
& \hat{h}_i^t := \frac{\sum_{\vz'_j \in \hat{\calZ}^{top_K}_i }\exp(\vz_i^\mathsf{T} \vz'_j)}{\sum_{\vz'_j \in \hat{\calZ}_i} \exp(\vz_i^\mathsf{T} \vz'_j)}, \\
& h_i^t := \rho h_i^{t-1}+(1-\rho) \hat{h}_i^t 
\label{eqn:headness}
\end{split}
\end{equation}
where $\hat{\calZ}_i = \calZ' \setminus \vz'_i$; $\hat{\calZ}^{top_K}_i$ denotes the set containing the $K\%$ most similar representations to $\vz_i$ in $\hat{\calZ}$; $\rho$ denotes a momentum hyperparameter with a value between 0 and 1.

Subsequently, we determine a temperature value $\tau_i$ for each instance $\mI_i$ using $h_i^t$. To avoid extreme values, we constrain $h_i^t$ to fall within the lowest 10\% ($h^{low}$) and highest 10\% ($h^{high}$) of the scores. We then apply min-max normalization to adjust the score to fall within the range between $\tau^{min}$ and $\tau^{max}$. Specifically, the temperature value $\tau_i$ for $\mI_i$ is calculated as follows:
\begin{equation}
\begin{split}
\tau_i := \frac{\bar{h}_i^t - \min(\calH^t)}{\max(\calH^t)-\min(\calH^t)}(\tau^{max} - \tau^{min})+\tau^{min} 
\label{eqn:temperature}
\end{split}
\end{equation}
where $\bar{h}_i^t := \min ( \max (h_i^t, h^{low}), h^{high} )$; $\calH^t$ represents the set containing the headness scores for all samples. For efficiency, $\tau_i$ and $h_i^t$ are updated at every epoch.

The two contrastive losses $\calL^u_{rep}$ and $\calL^s_{rep}$ in~\eref{eqn:cont_loss_prelim} are modified using the estimated instance-wise temperature value $\tau_i$ as follows:
\begin{equation}
\begin{split}
& \calL^u_{rep}:= -\log \frac{\exp(\vz_i^\mathsf{T} \vz'_i/\tau_i)}{\sum_{\vz'_j \in \hat{\calZ}_i } \exp(\vz_i^\mathsf{T} \vz'_j/\tau_i)}, \\
& \calL^s_{rep} := - \frac{1}{|\calZ^p_i|}\sum_{\vz'_k \in \calZ^p_i} \log \frac{\exp(\vz_i^\mathsf{T} \vz'_k/\tau_i)}{\sum_{\vz'_j \in \hat{\calZ}_i } \exp(\vz_i^\mathsf{T} \vz'_j/\tau_i)}_.
\label{eqn:loss_unsupervised}
\end{split}
\end{equation}

\vspace{1mm}
\noindent \textbf{Soft Attention Module}. 
Since $\calI^u_o$ and $\calI^l_o$ contain target objects along with background or adjacent objects, we investigate a soft attention module (SAM) to encode object-specific features. Although we generate pseudo-masks $\mM^u$ for $\calD^u$ using $f_o(\cdot)$, directly using these pseudo-masks to encode object-specific features is risky due to noisy boundaries. To address this, we train an efficient attention module using the pseudo-masks $\mM^u$ for $\calD^u$ and ground-truth masks $\mM^l$ for $\calD^l$. We integrate this attention module into every stage of the CNN backbone. Additionally, we utilize pooled feature maps and embedding functions with depth reduction to reduce computational complexity.


Given a feature map $\mF \in \mathbb{R}^{D \times \bar{H} \times \bar{W}}$ at the end of each stage in the backbone, we first reduce its dimensions to decrease subsequent computations. Specifically, we apply spatial average pooling to $\mF$ with varying receptive fields. Then, we use $M$ embedding functions $\eta_i(\cdot)$ to generate $M$ outputs $\mP_i \in \mathbb{R}^{d \times s_i \times s_i}$. Here, $i$ indexes the $M$ outputs, $d$ represents the reduced depth produced by the embedding functions, and $s_i \times s_i$ denotes the resulting spatial dimension after pooling. Subsequently, we reshape $\mP_i$ into $\hat{\mP}_i \in \mathbb{R}^{d \times s_i^2}$ and concatenate them to obtain $\bar{\mP} \in \mathbb{R}^{d \times (s_1^2 + s_2^2 + \cdots + s_M^2)}$.

Then, we compute a pairwise affinity matrix $\mA$ by projecting $\mF$ using an embedding function $\phi(\cdot)$ and multiplying the result by $\bar{\mP}$ (\ie $\mA := \bar{\mP}^T \phi(\mF)$). The matrix $\mA \in \mathbb{R}^{(s_1^2 + s_2^2 + \cdots + s_M^2) \times \bar{H} \times \bar{W}}$ represents the spatial relations between the pooled feature map $\mP$ and $\mF$. Additionally, we project $\mF$ using a function $\psi(\cdot)$ and apply global average pooling along the channel dimension, generating a map $\mG$.

Finally, we concatenate $\mA$ with $\mG$ and apply an embedding function $\nu(\cdot)$ followed by a sigmoid function to obtain an attention map $\mS \in \mathbb{R}^{1 \times \bar{H} \times \bar{W}}$. This map $\mS$ is then element-wise multiplied with each channel of $\mF$ to produce the output $\mO \in \mathbb{R}^{D \times \bar{H} \times \bar{W}}$ of the attention module.
\begin{equation}
\begin{split}
\mO := \mS \odot \mF := \sigma( \nu ( [\mA, \mG] ) ) \odot \mF
\label{eqn:soft_attention}
\end{split}
\end{equation}
where $\odot$ and $\sigma(\cdot)$ denote element-wise multiplication and the sigmoid function, respectively; $[\cdot, \cdot]$ represents concatenation. Each of the embedding functions ($\eta(\cdot)$, $\psi(\cdot)$, $\phi(\cdot)$) consists of a $1 \times 1$ convolution layer, batch normalization, and ReLU activation. In comparison, $\nu(\cdot)$ contains only a $1 \times 1$ convolution layer and batch normalization.


To train the soft attention modules, we use object masks $\mM^u$ from $f_o(\cdot)$ for $\calD^u$ and ground-truth masks $\mM^l$ for $\calD^l$. Because the pseudo-masks for $\calD^u$ are noisy, especially near object boundaries~\cite{wang2022noisy}, we utilize a weight map $\mW$ to reduce reliance on these regions. Specifically, the attention loss $\calL_{att}$ is computed as follows:
\begin{equation}
\begin{split}
& \calL_{att} := \frac{1}{HW} \sum_{i=1}^H \sum_{j=1}^W \mW_{ij} \norm{\mS_{ij}-\mM_{ij}}^2_2, \\
\end{split}
\end{equation}
where $\mW_{ij}$ is set to $w$ if $d_{ij} \leq \bar{d}$ and $\mM \in \{\mM^u\}$, and to 1 otherwise. Here, $d_{ij}$ denotes the Euclidean distance from ($i, j$) to the nearest object boundary; $\bar{d}$ is a hyperparameter that defines the boundary regions; and $w$ is a weighting coefficient ($\leq 1$). Since SAM is applied after each stage of the backbone, $\calL_{att}$ is obtained by averaging all the corresponding losses.


\begin{figure}[!t] 
\begin{minipage}{0.24\linewidth}
\centerline{\includegraphics[width=0.97\linewidth,height=0.06\textheight]{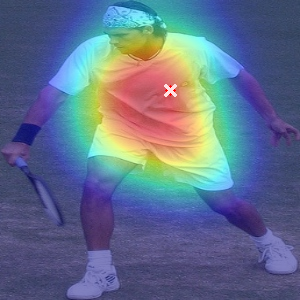}}
\end{minipage}
\begin{minipage}{0.24\linewidth}
\centerline{\includegraphics[width=0.97\linewidth,height=0.06\textheight]{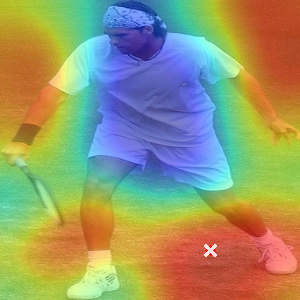}}
\end{minipage}
\hspace{0.1mm}
\begin{minipage}{0.24\linewidth}
\centerline{\includegraphics[width=0.97\linewidth,height=0.06\textheight]{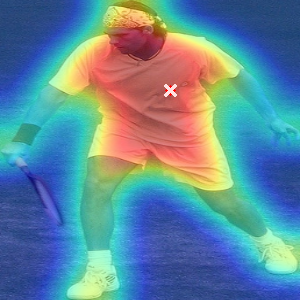}}
\end{minipage}
\begin{minipage}{0.24\linewidth}
\centerline{\includegraphics[width=0.97\linewidth,height=0.06\textheight]{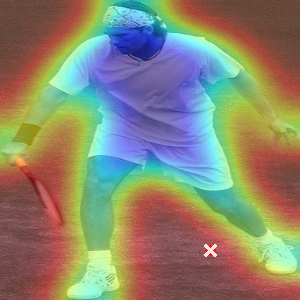}}
\end{minipage}

\begin{minipage}{0.486\linewidth}
\centerline{\footnotesize (a) RGA-S~\cite{zhang2020relation}}
\end{minipage}
\begin{minipage}{0.486\linewidth}
\centerline{\footnotesize (b) Ours (SAM)}
\end{minipage}
\caption{Visualization of the pairwise affinity between the white cross marked location and other pixels.}
\label{fig:SAM}
\end{figure}

\fref{fig:SAM} visualizes the pairwise affinity between a marked position and other pixels. The results show that our method, which uses pooled feature maps, is more robust than the previous approach~\cite{zhang2020relation}.

\vspace{1mm}
\noindent \textbf{Deep Clustering for GCD}. 
To avoid the separate semi-supervised clustering step in~\cite{vaze2022generalized}, we employ a deep clustering method similar to those in~\cite{fomenko2022learning, wen2023parametric}. However, unlike~\cite{fomenko2022learning}, our method does not rely on an experimentally selected target prior distribution. Additionally, it is designed to handle imbalanced data, in contrast to~\cite{wen2023parametric}. Specifically, we use the method from~\cite{Zhang2021Supporting} for clustering $\calI^u_o$, with a minor modification: replacing L2 distance with cosine similarity.

We compute the KL-divergence-based loss $\calL^u_{cls}$ on $\calI^u_o$ for unsupervised clustering in~\cite{Zhang2021Supporting}, as follows:
\begin{equation}
\begin{split}
\calL^u_{cls} := \sum_{c=1}^C \bar{p}_{ic} \log \frac{\bar{p}_{ic}}{q_{ic}}
\label{eqn:loss_KL_divergence}
\end{split}
\end{equation}
where $q_{ic}$ is the probability of $\mI_i$ belonging to class $c$; $\bar{p}_{ic}$ is the auxiliary target class probability; and $C$ is the total number of clusters/classes. 

Independent from deep clustering, we additionally compute the typical cross-entropy loss $\calL^s_{cls}$ on $\calD^l$ for supervised classification, as follows:
\begin{equation}
\begin{split}
\calL^s_{cls} := \sum_{c=1}^C - y_{ic} \log q_{ic}
\label{eqn:loss_cross_entropy}
\end{split}
\end{equation}
where $\vy_i$ denotes the one-hot encoded label for $\mI_i$.

\vspace{1mm}
\noindent \textbf{Total Loss for GCD}. 
The total loss $\calL_{gcd}$ for $f_d(\cdot)$ is computed as the weighted sum of the two contrastive losses, the two classification losses, and the attention loss, as follows:
\begin{equation}
\calL_{gcd} := \calL_{att} + (1-\lambda) \calL^u_{rep} + \lambda \calL^s_{rep} + (1-\lambda) \calL^u_{cls} + \lambda \calL^s_{cls}
\label{eqn:gcd_loss}
\end{equation}
where $\lambda$ is a hyperparameter used to balance the losses, following~\cite{wen2023parametric}.



\subsection{Reliability-Based Dynamic Learning}
\label{sec:dynamic_learning}
We generate pseudo-masks $\mM^u$ and pseudo-class labels $\vy^u$ for $\calD^u$ using the method described in~\sref{sec:class_discovery}. Subsequently, we train an instance segmentation network $f_s(\cdot)$ that can segment instances of both known and novel classes using $\calD^l$ and $\calD^u$ with pseudo-labels. To address the issues of inaccurate pseudo-labels and imbalanced instance distributions across classes, we propose a reliability-based dynamic learning (RDL) method. It applies different reliability criteria to each class to avoid excluding all samples from tail classes. Additionally, it adjusts these criteria during training to use diverse data in the early stages while relying only on reliable pseudo-labels in the later stages.

Inspired by~\cite{yang2022st++}, we use holistic stability to measure the reliability of the pseudo-labels. At every fixed number of epochs during training $f_d(\cdot)$, we save the model at that point in time. We then apply these saved models to object images $\mI_i \in \calI^u_o$ to compute the probability $q^{\bar{t}}_{ic}$ of $\mI_i$ belonging to class $c$, where $\bar{t}$ denotes the index of the stored models, ranging from 1 to $\bar{T}$. Subsequently, we compute the stability $s_{i}$ of the probabilities by comparing $q^{\bar{T}}_{ic}$ from the final model with $q^{\bar{t}}_{ic}$ from the intermediate models, as follows:
\begin{equation}
s_{i} = \sum_{\bar{t}=1}^{\bar{T} - 1} \frac{1}{KL(\vq^{\bar{T}}_{i} || \vq^{\bar{t}}_{i})}
\label{eqn:stability}
\end{equation}
where $KL(\cdot||\cdot)$ represents the Kullback-Leibler divergence, and $\vq^{\bar{t}}_{i} \in \mathbb{R}^{|\calC|}$. Since a higher $s_{i}$ indicates greater stability, \cite{yang2022st++} considered pseudo-labels with the lowest $r\%$ scores unreliable.


However, applying the same criteria to all data may result in categorizing most samples of a certain class as unreliable. Specifically, for imbalanced data, instances of tail classes tend to have lower $s_{i}$ than those of head classes due to the smaller number of training samples. Additionally, because neural networks tend to first memorize easy samples and then gradually learn harder instances during training~\cite{arpit2017closer}, difficult samples/classes often have lower $s_{i}$ than easier ones. To address these issues, we propose to use class-wise reliability criteria, which consider pseudo-labels with the lowest $r\%$ scores per class as unreliable.



Additionally, we gradually increase the portion $r\%$ of unreliable samples to initially learn from all data and later optimize using only reliable samples. The idea is that, in the early stages, having a larger number of diverse samples is more important than the accuracy of pseudo-labels, while pseudo-label quality becomes more crucial in later stages. Specifically, we first compute $\bar{t}_i$ for each instance $i$ by finding the rank of $s_i$ among the lowest values within its class. We assign $\gamma$ to $\bar{t}_i$ if its proportional rank falls between $\frac{\gamma}{T_{is}}$ and $\frac{\gamma+1}{T_{is}}$. Then, the reliability-based adjustment weight $\kappa^t_i$ is computed as follows:
\begin{equation}
\kappa^t_i = \sqrt{1-\Big( \max \Big(\frac{ t - \bar{t}_i}{T_{is}}, 0 \Big) \Big)^2} \\
\label{eqn:adjusting_weight}
\end{equation}
where $t$ and $T_{is}$ denote the current epoch and the total number of epochs for training $f_s(\cdot)$, respectively.


%

Finally, we employ SOLOv2~\cite{wang2020solov2} for $f_s(\cdot)$ and use its loss function with modifications for training. First, we replace the focal loss~\cite{lin2017focal} with the equalized focal loss~\cite{li2022equalized}, which performs better on imbalanced data. We use this modified loss $\calL_{s}$ for $\calD^l$ and this loss multiplied by $\kappa^t_i$ for $\calD^u$. Therefore, the total instance segmentation loss $\calL_{is}$ is computed as follows:
\begin{equation}
\begin{split}
\calL_{is} = \sum_{i \in \calB^l} \calL_{s}(\mI^l_i,\mM^l_i,\vy^l_i) + \sum_{i \in \calB^u} \kappa^t_i \calL_{s}(\mI^u_i,\mM^u_i,\vy^u_i)
\label{eqn:loss_instance_segmentation}
\end{split}
\end{equation}
where $\calB^l$ and $\calB^u$ denote the sets containing the indices of the data from $\calD^l$ and $\calD^u$, respectively.




\section{Experiments and Results}
\label{sec:result}
\subsection{Experimental Setting}
\noindent \textbf{Dataset}.
We conducted experiments in two settings: COCO$_{half}$ + LVIS and LVIS + VG, following~\cite{fomenko2022learning}. In the COCO$_{half}$ + LVIS setting, we consider 80 COCO classes as known classes and aim to discover 1,123 disjoint LVIS classes from the total 1,203 LVIS classes. Among the 100K training images in the LVIS dataset~\cite{gupta2019lvis}, we use 50K images with labels for the 80 COCO classes for $\calD^l$ and the entire 100K images without labels for $\calD^u$. We use the 20K LVIS validation images for evaluation.

In the LVIS + VG setting, we utilize 1,203 LVIS classes as known classes and aim to discover 2,726 disjoint classes in the Visual Genome (VG) v1.4 dataset~\cite{krishna2017visual}. We use the entire 100K LVIS training data for $\calD^l$ and the combined 158K LVIS and VG training images for $\calD^u$. For evaluation, we use 8K images that appear in both the LVIS and VG validation sets. Although the VG dataset contains over 7K classes, only 3,367 classes appear in both $\calD^u$ and the validation data. After excluding the 641 classes that overlap with the known classes, we aim to discover 2,726 classes.


\vspace{1mm}
\noindent \textbf{Implementation Details}. 
We use ResNet-50 as the backbone for both $f_d(\cdot)$ and $f_s(\cdot)$. For $f_d(\cdot)$, we apply the SAM to all four stages of the backbone. We train $f_d(\cdot)$ for 390 epochs with $K=1\%$, $\tau^{min}=0.07$, $\tau^{max}=1$, and $\lambda=0.35$. For the SAM, we set $M=3$, $d=D/8$, $w=0.25 \cdot stg$, $\hat{d}=1$, $s_1=\lfloor 18/(2^{stg-1}) \rceil$, $s_2=\lfloor 12/(2^{stg-1}) \rceil$, and $s_3=\lfloor 8/(2^{stg-1}) \rceil$, where $stg$ is the stage index and $\lfloor \cdot \rceil$ denotes rounding. We train $f_s(\cdot)$ for 36 epochs with $\bar{T}=3$. The experiments were conducted on a computer with two Nvidia GeForce RTX 3090 GPUs, an Intel Core i9-10940X CPU, and 128 GB RAM. The code will be publicly available on GitHub upon publication to ensure reproducibility.


\vspace{1mm}
\noindent \textbf{Evaluation Metric}. 
We first apply the Hungarian algorithm~\cite{kuhn1955hungarian} to find a one-to-one mapping between the discovered classes and the ground-truth novel classes, following~\cite{fomenko2022learning}. We then calculate $\text{mAP}_{.50:.05:.95}$ for all, known, and novel classes. $\text{mAP}_{.50:.05:.95}$ is computed using mask labels for the COCO$_{half}$ + LVIS setting and using bounding box labels for the LVIS + VG setting due to the absence of mask labels in the VG dataset.

\subsection{Result}
Tables~\ref{tab:result_LVIS} and~\ref{tab:result_VG} present quantitative comparisons for the COCO$_{half}$ + LVIS and LVIS + VG settings, respectively. We compare the proposed method with the $k$-means~\cite{macqueen1967some} baseline and previous works including ORCA~\cite{cao2021open}, UNO~\cite{fini2021unified}, SimGCD~\cite{wen2023parametric}, and RNCDL~\cite{fomenko2022learning}. Additionally, we report results obtained by replacing the GCD model in our framework with recent GCD methods: $\mu$GCD~\cite{Vaze2023No} and NCDLR~\cite{zhang2023novel}. The results demonstrate that our method significantly outperforms the previous state-of-the-art method~\cite{fomenko2022learning} as well as methods focusing on balanced and curated datasets~\cite{cao2021open, fini2021unified, wen2023parametric}, across all metrics and settings. ``RNCDL w/ ran. init.'' refers to the method with random initialization in~\cite{fomenko2022learning}. 

\begin{table}[!t]
\begin{minipage}{1\linewidth}
\small
\begin{tabular}{ >{\centering}m{0.33\textwidth}| >{\centering}m{0.15\textwidth}| >{\centering}m{0.15\textwidth}| >{\centering\arraybackslash}m{0.15\textwidth} } 
\toprule
 $\text{Method}$   &  $\text{mAP}_{\text{all}}$ & $\text{mAP}_{\text{known}}$ & $\text{mAP}_{\text{novel}}$ \\ 
\midrule
$k$-means & 1.48  & 17.24 & 0.13 \\
RNCDL w/ ran. init. & 1.87  & 22.85 & 0.09 \\
ORCA  				& 3.19  & 21.61 & 1.63 \\
UNO  				& 3.42  & 22.34 & 1.86  \\  
SimGCD 				& 4.06  & 23.91 & 2.47  \\ 
RNCDL 				& 6.69  & 25.21 & 5.16  \\  
$\mu$GCD$^\dagger$  & 4.93 & 25.36 & 3.58   \\ 
NCDLR$^\dagger$ 	& 9.42 & 27.81 & 8.06   \\ 
Ours & 		\textbf{12.85}& \textbf{35.57}& \textbf{11.24}   \\ 
\bottomrule
\end{tabular}
\end{minipage}
\caption{Comparison for the COCO$_{half}$ + LVIS setting. $^\dagger$ indicates that only the GCD model in our method is replaced by the corresponding GCD method while all other components of our method remain unchanged.}
\label{tab:result_LVIS}
\end{table}

\begin{table}[!t]
\begin{minipage}{1\linewidth}
\small
\centering
\begin{tabular}{ >{\centering}m{0.33\textwidth}| >{\centering}m{0.15\textwidth}| >{\centering}m{0.15\textwidth}| >{\centering\arraybackslash}m{0.15\textwidth} } 
\toprule
 $\text{Method}$   &  $\text{mAP}_{\text{all}}$ & $\text{mAP}_{\text{known}}$ & $\text{mAP}_{\text{novel}}$ \\ 
\midrule
RNCDL w/ ran. init. & 2.13   & 7.71 & 0.81 \\
SimGCD 				& 2.37  & 9.13 & 0.92	\\
RNCDL 				& 4.46   & 12.55 & 2.56 \\  
$\mu$GCD$^\dagger$  & 2.64 & 9.74 & 0.98   \\ 
NCDLR$^\dagger$ 	& 4.71 & 12.87 & 2.63   \\ 
Ours 				& \textbf{5.21}& \textbf{13.28}& \textbf{3.27}  \\  
\bottomrule
\end{tabular}
\end{minipage}
\caption{Quantitative comparison for the LVIS + VG setting.}
\label{tab:result_VG}
\end{table}

\begin{figure}[!t] 
\begin{minipage}{0.49\linewidth}
\centerline{\includegraphics[width=0.97\linewidth,height=0.1\textheight]{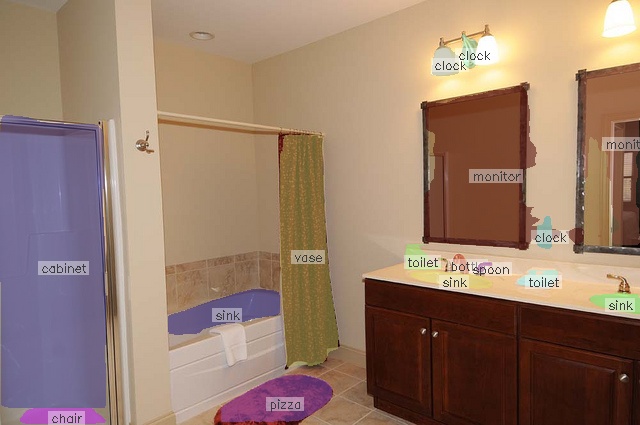}}
\end{minipage}
\begin{minipage}{0.49\linewidth}
\centerline{\includegraphics[width=0.97\linewidth,height=0.1\textheight]{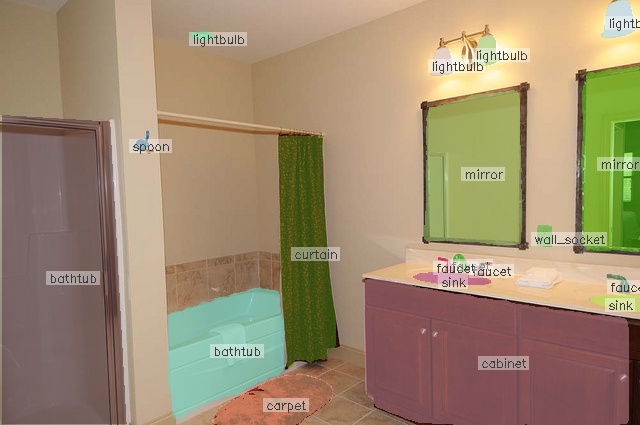}}
\end{minipage}

\begin{minipage}{0.49\linewidth}
\centerline{\footnotesize (a) RNCDL}
\end{minipage}
\begin{minipage}{0.49\linewidth}
\centerline{\footnotesize (b) Ours}
\end{minipage}

\begin{minipage}{0.49\linewidth}
\centerline{\includegraphics[width=0.97\linewidth,height=0.1\textheight]{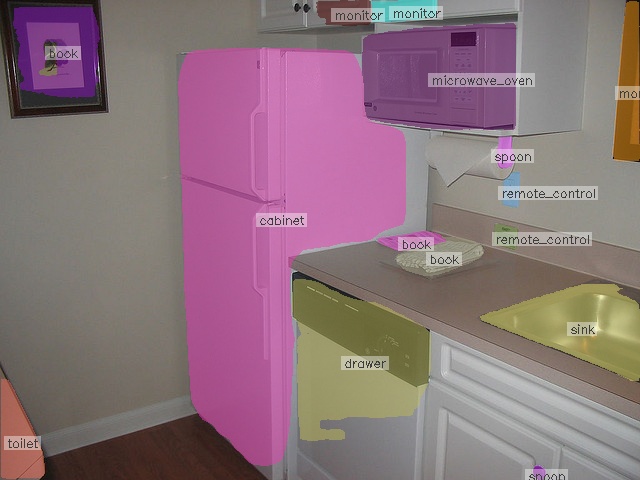}}
\end{minipage}
\begin{minipage}{0.49\linewidth}
\centerline{\includegraphics[width=0.97\linewidth,height=0.1\textheight]{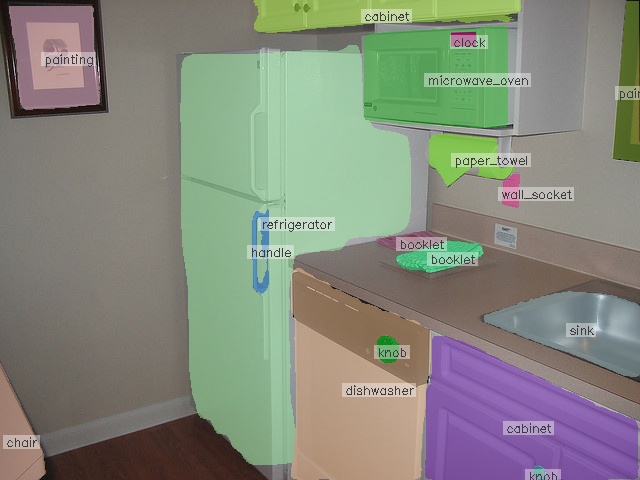}}
\end{minipage}

\begin{minipage}{0.49\linewidth}
\centerline{\footnotesize (c) RNCDL}
\end{minipage}
\begin{minipage}{0.49\linewidth}
\centerline{\footnotesize (d) Ours}
\end{minipage}
\caption{Qualitative results. (a) and (b) are COCO$_{half}$ + LVIS setting; (c) and (d) are LVIS + VG setting.}
\label{fig:result}
\end{figure}

\fref{fig:result} shows qualitative comparisons with RNCDL~\cite{fomenko2022learning} for both settings. 



\subsection{Analysis}
We conducted all the ablation studies and analyses using the COCO$_{half}$ + LVIS setting unless otherwise noted. \tref{tab:ablation} presents an ablation study of the proposed components. The baseline model is constructed by using a fixed value for the temperature parameters for all samples in~\eref{eqn:loss_unsupervised}, excluding SAM, and treating all pseudo-labels as equivalent to human annotations. The results demonstrate that each module provides additional improvements across all metrics.


\begin{table}[!t]
\begin{minipage}{1\linewidth}
\small
\centering
\begin{tabular}{ >{}m{0.33\textwidth}| >{\centering}m{0.15\textwidth}| >{\centering}m{0.15\textwidth}| >{\centering\arraybackslash}m{0.15\textwidth} } 
\toprule
Method &  $\text{mAP}_{\text{all}}$ & $\text{mAP}_{\text{known}}$ & $\text{mAP}_{\text{novel}}$  \\
\midrule
Baseline & 5.79 & 24.96 & 4.19 \\
+ ITA    & 7.64 & 26.84 & 6.09 \\
+ ITA + SAM   & 10.71 & 31.62 & 9.08 \\
+ ITA + SAM + RDL &\textbf{12.85}& \textbf{35.57}& \textbf{11.24} \\
\bottomrule
\end{tabular}
\end{minipage}
\caption{Ablation study on the proposed components.}
\label{tab:ablation}
\end{table}

\begin{table}[!t]
\begin{minipage}{1\linewidth}
\small
\centering
\begin{tabular}{ >{\centering}m{0.19\textwidth}| >{\centering}m{0.17\textwidth}| >{\centering}m{0.11\textwidth}| >{\centering}m{0.13\textwidth}| >{\centering\arraybackslash}m{0.13\textwidth} } 
\toprule
$\tau^{min}$, $\tau^{max}$   & Method &  $\text{mAP}_{\text{all}}$ & $\text{mAP}_{\text{known}}$ & $\text{mAP}_{\text{novel}}$  \\
\midrule 
\multirow{2}{*}{$0.07$, $0.5$} & TS   & 10.86 & 33.72 & 9.19 \\
 				& Ours (ITA) & 11.67 & 34.63 & 10.01 \\
\midrule
\multirow{2}{*}{$0.07$, $1$} & TS   & 11.43 & 34.25 &9.79\\
		 		& Ours (ITA) &\textbf{12.85} & \textbf{35.57} & \textbf{11.24} \\
\bottomrule
\end{tabular}
\end{minipage}
\caption{Comparison on temperature assignment methods.}
\label{tab:analysis_temperature}
\end{table}

\tref{tab:analysis_temperature} compares the proposed ITA method with the TS method from~\cite{kukleva2023temperature}. Although the TS method is also designed for imbalanced data, our ITA method consistently outperforms it for both the selected hyperparameters ($0.07$, $1$) and an alternative set ($0.07$, $0.5$). Additionally, the table shows that the chosen hyperparameters yield better performance than the alternative set.

\tref{tab:analysis_attention} compares the proposed SAM with previous attention methods, including MGCAM~\cite{song2018mask}, MGFPM~\cite{wang2021mask}, CBAM-S~\cite{woo2018cbam}, and RGA-S~\cite{zhang2020relation}. The results for the other methods are obtained by replacing only the SAM in the proposed method with the corresponding methods. The results demonstrate that our SAM outperforms all other methods across all metrics.

\begin{table}[!t]
\begin{minipage}{1\linewidth}
\small
\centering
\begin{tabular}{ >{\centering}m{0.3\textwidth}| >{\centering}m{0.16\textwidth}| >{\centering}m{0.16\textwidth}| >{\centering\arraybackslash}m{0.16\textwidth} } 
\toprule
 $\text{Method}$ & $\text{mAP}_{\text{all}}$ & $\text{mAP}_{\text{known}}$ & $\text{mAP}_{\text{novel}}$ \\ 
\midrule
MGCAM & 10.34  & 29.37 & 8.79  \\
MGFPM & 10.63  & 31.15 & 9.01   \\  
CBAM-S & 11.19   & 31.83 & 9.56  \\
RGA-S & 11.62   & 33.48 & 9.98 \\  
\midrule
Ours (SAM) & \textbf{12.85}& \textbf{35.57}& \textbf{11.24} \\  
\bottomrule
\end{tabular}
\end{minipage}
\caption{Comparison with other attention methods.}
\label{tab:analysis_attention}
\end{table}

\begin{table}[!t]
\begin{minipage}{1\linewidth}
\small
\centering
\begin{tabular}{ >{\centering}m{0.3\textwidth}| >{\centering}m{0.16\textwidth}| >{\centering}m{0.16\textwidth}| >{\centering\arraybackslash}m{0.16\textwidth} } 
\toprule
$\text{Method}$ & $\text{mAP}_{\text{all}}$ & $\text{mAP}_{\text{known}}$ & $\text{mAP}_{\text{novel}}$ \\ 
\midrule
$r=25$\% & 11.56   & 32.45 & 9.92 \\  
$r=50$\% & 11.79   & 33.36 & 10.17 \\  
$r=75$\% & 11.63   & 32.79 & 10.01 \\  
\midrule
Ours (RDL) &\textbf{12.85}& \textbf{35.57}& \textbf{11.24} \\  
\bottomrule
\end{tabular}
\end{minipage}
\caption{Comparison of reliability-based dynamic learning with fixed criteria.}
\label{tab:analysis_reliability}
\end{table}

\tref{tab:analysis_reliability} compares our RDL method with the fixed and global reliability criteria in ST++~\cite{yang2020superpixel}. The results confirm the importance of using selected pseudo-labels based on adjusted reliability criteria for each class throughout the training process.

\begin{table}[!t]
\begin{minipage}{1\linewidth}
\small
\centering
\begin{tabular}{ >{\centering}m{0.3\textwidth}| >{\centering}m{0.16\textwidth}| >{\centering}m{0.16\textwidth}| >{\centering\arraybackslash}m{0.16\textwidth} } 
\toprule
 $\text{Method}$ & $\text{mAP}_{\text{all}}$ & $\text{mAP}_{\text{known}}$ & $\text{mAP}_{\text{novel}}$ \\ 
\midrule
Mask R-CNN  & 11.82  & 32.74 & 10.13  \\
OLN  & 12.07  & 33.56 & 10.52   \\  
LDET  & 12.36 & 34.63  & 10.74  \\
UDOS  & 12.54 & 35.12  & 11.05 \\  
GGN  & \textbf{12.85} & \textbf{35.57} & \textbf{11.24} \\  
\bottomrule
\end{tabular}
\end{minipage}
\caption{Comparison of class-agnostic instance mask generation methods.}
\label{tab:analysis_open_world}
\end{table}

\tref{tab:analysis_open_world} presents the results of our method using various class-agnostic instance mask generation models, including Mask R-CNN~\cite{he2017mask}, OLN~\cite{Kim2022Learning}, LDET~\cite{Saito2022Learning}, UDOS~\cite{kalluri2023open}, and GGN~\cite{wang2022open}. The results demonstrate that our method consistently outperforms the previous state-of-the-art method~\cite{fomenko2022learning}, regardless of the choice of mask generation model.


\section{Conclusion}
Towards open-world instance segmentation, we present a novel GCD method in instance segmentation. To address the imbalanced distribution of instances, we introduce the instance-wise temperature assignment method as well as class-wise and dynamic reliability criteria. The former aims to improve the embedding space for class discovery, and the criteria are designed to effectively utilize pseudo-labels from the GCD model. Additionally, we propose an efficient soft attention module. The experimental results in two settings demonstrate that the proposed method outperforms previous methods by effectively discovering novel classes and segmenting instances of both known and novel categories.

Regarding limitations, this work assumes the full availability of labeled and unlabeled datasets from the beginning. Thus, it is suboptimal for scenarios where data is provided sequentially, such as in robot navigation. Additionally, we assume prior knowledge of the total number of classes, following most previous works.

\section*{Acknowledgments}
This work was supported by the National Research Foundation of Korea(NRF) grant funded by the Korea government(MSIT) (No. RS-2023-00252434). 

\bibliography{aaai25}

\end{document}